\documentclass[letterpaper, 10 pt, conference]{ieeeconf}  

\IEEEoverridecommandlockouts                              

\usepackage{verbatim}

\usepackage[ruled, linesnumbered]{algorithm2e}

\usepackage{cite}

 \usepackage{todonotes}

\title{\LARGE \bf
Surface Following using Deep Reinforcement Learning and a GelSight Tactile Sensor
}
\author{Chen Lu$^{1}$, Jing Wang$^{1,2}$ and Shan Luo$^{1}$ 
\thanks{$^{1}$smARTLab, Department of Computer Science, University of Liverpool, Liverpool L69 3BX, UK.}%
\thanks{$^{2}$Northwest Institute of Mechanical and Electrical Engineering, Xianyang 712099, China}%
\thanks{Emails: \{c.lu15, jing.wang4, shan.luo\}@liverpool.ac.uk}
}

\usepackage{ifpdf}
\usepackage{graphicx}
\ifpdf

\graphicspath{{Figs/}{Figs/PDF/}}
\else
\graphicspath{{Figs/}}
\fi

\usepackage{hhline}

\usepackage{calc}
\let\labelindent\relax
\usepackage{enumitem}

\usepackage{hyperref}
\hypersetup{
	colorlinks = true,
	linkcolor = {blue},
	citecolor = {blue}
}
\usepackage{url}

\usepackage{nomencl}
\usepackage{multirow}

\usepackage{subcaption}

\usepackage{array}
\usepackage{booktabs}

\newcolumntype{P}[1]{>{\centering\arraybackslash}p{#1}}

\usepackage[flushleft]{threeparttable}

\usepackage{cite}

\usepackage{times}
\usepackage{epsfig}
\usepackage{graphicx}
\usepackage{amsmath}
\usepackage{amssymb}

\begin{document}

\maketitle
\thispagestyle{empty}
\pagestyle{empty}

\begin{abstract}
Tactile sensors can provide detailed contact information that can facilitate robots to perform dexterous, in-hand manipulation tasks. One of the primitive but important tasks is surface following that is a key feature for robots while exploring unknown environments or workspace of inaccurate modeling. In this paper, we propose a novel end-to-end learning strategy, by directly mapping the raw tactile data acquired from a GelSight tactile sensor to the motion of the robot end-effector. 
Experiments on a KUKA youBot platform equipped with the GelSight sensor show that 80\% of the actions generated by a fully trained
SFDQN model are proper surface following actions; the autonomous surface following test also indicates that the proposed solution works well on a test surface. 

\end{abstract}

\section{Introduction}
In many robot tasks, the robotic manipulator is required to contact and follow a surface stably. For example, when robots perform tasks like buffing, grinding, polishing and painting, it is necessary to maintain good contact between the manipulator end-effector and the surface of the object, and requires the end-effector to follow the object surface precisely, defined as a \textit{surface following} \cite{demey1994enhancing} or surface tracking \cite{araujo19963d} problem. It is also an important feature when robots explore environments that are unknown or of inaccurate modeling. Similar problems include contour following \cite{koch2011predictive} and surface exploration \cite{back2014control}. The former focuses more on exploring the shape of objects, while the latter aims to understand the physical properties of unknown objects through exploration, such as surface roughness, object shape and compliance. In this paper, we focus on the surface following problem whereas the methodologies developed can be extended to tackle the other two problems.



To achieve surface following, a robot manipulator needs to maintain constant and uniform contact with the object surface. This requires the robot to have the ability to sense and recognize its contact state with the surface, and the ability to control the manipulator to make real-time adjustments according to the surface variations. Therefore, the surface following task involves robot sensing, learning and control, which makes it a complicated problem. The common strategy to solve the surface following problem is to learn the desired trajectory and then control the robot motion according to the learned desired trajectory \cite{demey1994enhancing,koch2011predictive}. The obvious problem with this strategy is that the trajectory needs to be relearned for each object with new surface. In addition, camera vision has been used to sense the object surface and hence visual servoing can be applied to surface following \cite{chang2004cartesian}. However, vision cannot provide detailed information of physical properties, such as surface roughness and compliance, which can greatly affect the contact between robotic manipulator and object surface. To overcome this information loss, the haptic feedback can be used to perceive these physical properties and achieve more consistent surface following \cite{back2014control}.

\begin{figure}
 	\centering
 	\begin{subfigure}[b]{0.23\textwidth}
 		\centering
 		\includegraphics[height=4cm,width=4cm]{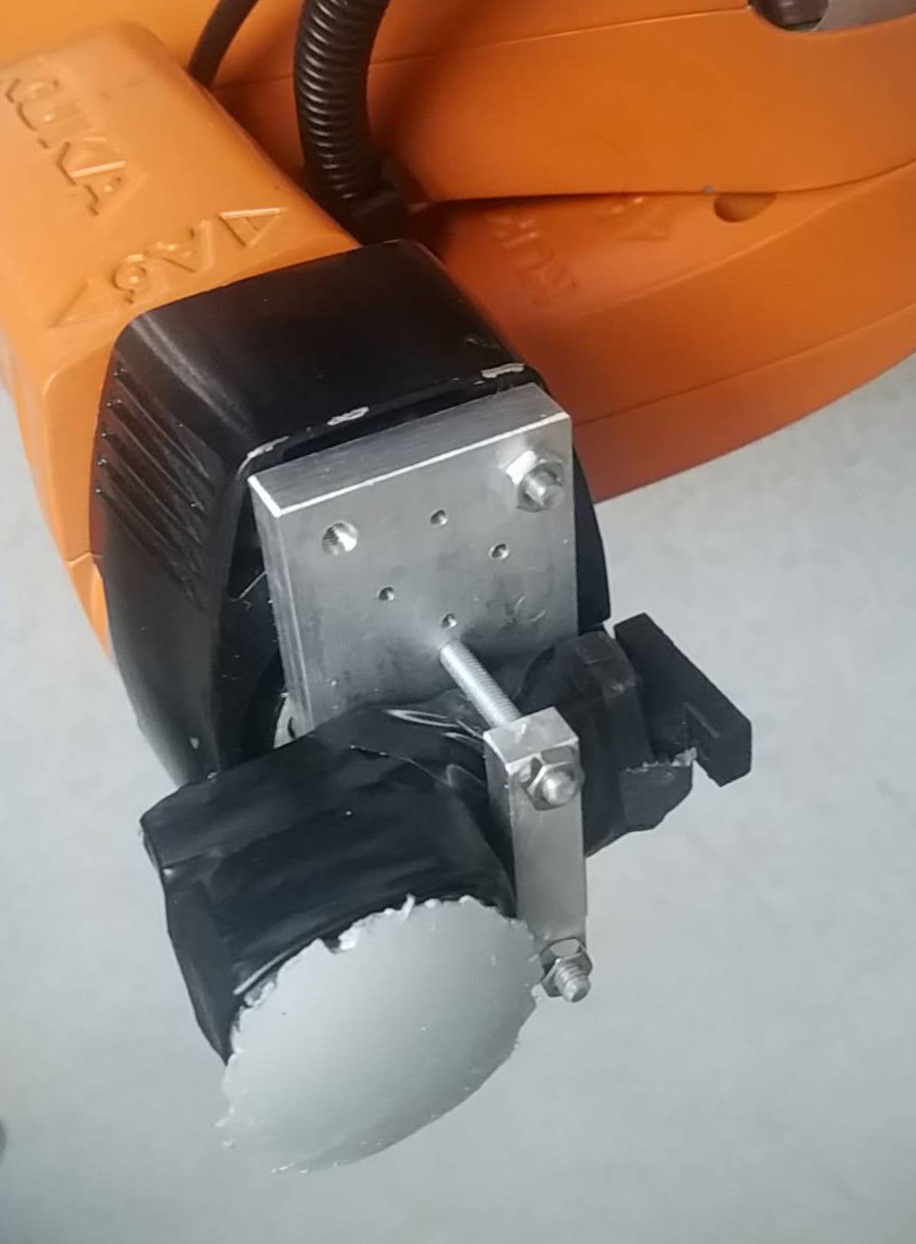}
 		\caption{}
 		\label{Fig:GelSightSensorEquipped}
 	\end{subfigure}%
 	\hfill
 	\begin{subfigure}[b]{0.23\textwidth}
 		\centering
 		\includegraphics[height=4cm,width=4cm]{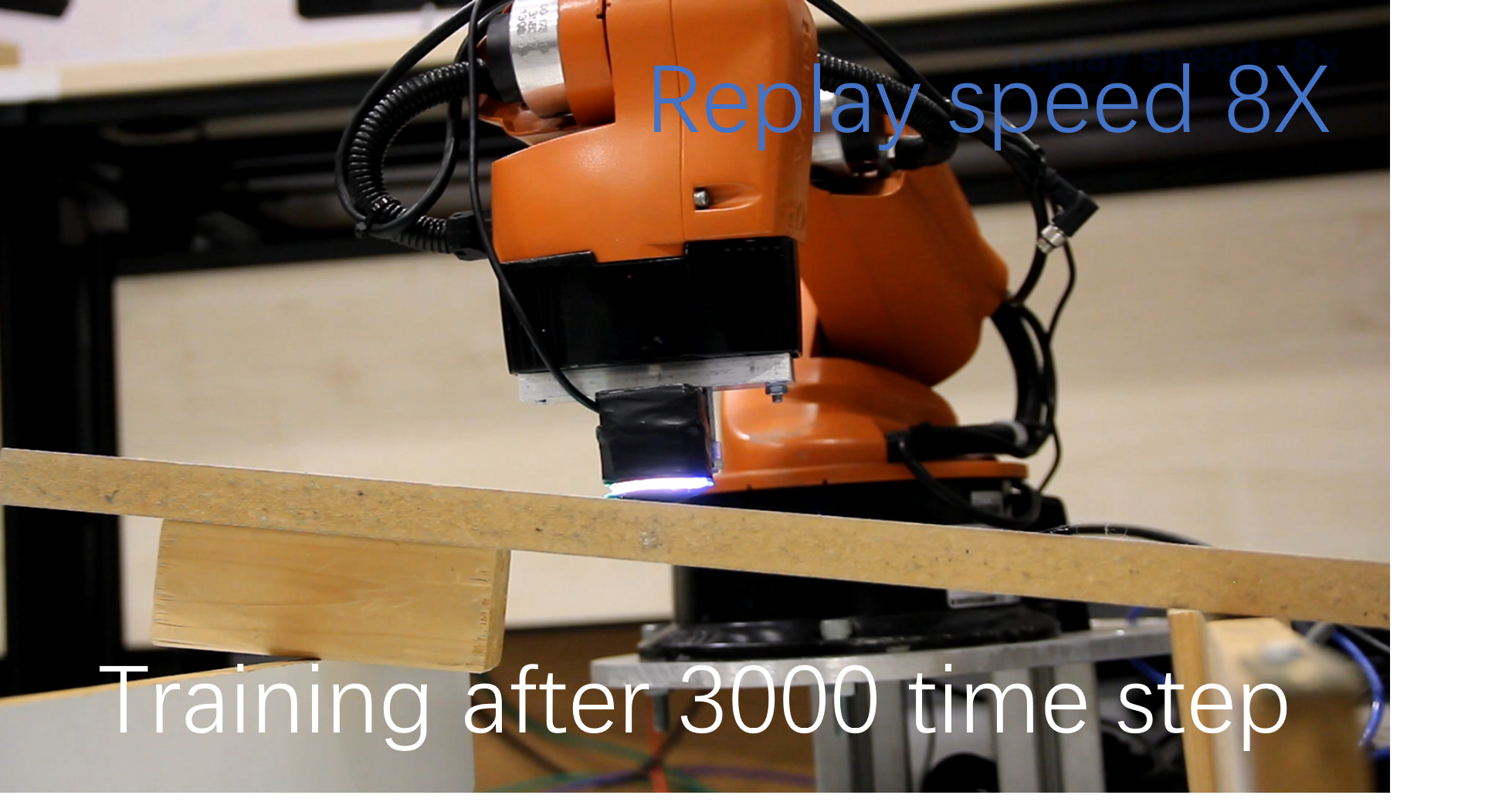}
 		\caption{}
 		\label{Fig:GelSightDoingSurfaceFollowing}
 	\end{subfigure}%

 	\caption{(a) The GelSight tactile sensor. (b) Robot performing surface following using a GelSight sensor.} 
 	\label{Fig:GelSight}
 \end{figure}
 
The main procedure of the existing strategies for surface following using haptic feedback can be summarized in two steps: The handcrafted features are first extracted from the haptic data, such as surface normal, surface tangent and contact force; the learned features are then used for controlling the end-effector movements. Due to error accumulation in the two-step strategy, it has high demand on the accuracy of sensors, robot hardware and control algorithms in order to achieve good performance. To overcome the challenge, in this paper, we propose a novel end-to-end learning strategy for surface following, by directly mapping raw tactile data to the end-effector control policies using deep reinforcement learning. We aim to maintain the contact area between the robot end-effector and the object surface into a fixed range while following the surface which is adaptable to various object surfaces. A GelSight tactile sensor \cite{johnson2009retrographic} is used to provide tactile feedback for facilitating the motion of the manipulator end-effector to follow different object surfaces. As shown in Fig.~\ref{Fig:GelSight}, the robot performs surface following using the learned policies with tactile feedback from the GelSight sensor equipped to a robot manipulator.

The rest of paper is organized as follows: Section~\ref{sec:RelatedWorks} reviews the related works; Section~\ref{sec:Methodologies} introduces the proposed end-to-end surface following algorithm; Section~\ref{sec:ExperimentSetup} illustrates and details the experiment setup; experimental results and analysis are given in Section~\ref{sec:ExperimentsResultsandAnalysis}; finally Section~\ref{sec:Conclusions} concludes the paper and points future directions.

\section{Related works}
\label{sec:RelatedWorks}

\subsection{Control Strategy for Surface Following}
There are mainly two kinds of control strategy for surface following in the literature. 
One is based on the trajectory control along the modelled surface, 
{the other is based on the motion control using feedback information while interacting with environment.}
For the trajectory control based strategy, the surface of the object is modelled first and then the trajectory of the end-point is planned according to it. The surface following performance of this strategy depends on the accuracy of the trajectory and the error between actual and modelled surfaces \cite{demey1994enhancing, koch2011predictive}. As the object surface needs to be modeled in advance to predict the trajectory, this strategy cannot be used to follow an unknown object.
{For the motion control based strategy, it focuses on the control of robot motion to maintain the desired contact status using real-time sensor information that obtained by interacting with object surface. In \cite{back2014control}, it uses sensors to obtain the information about surface normal and contact force, and then compute the surface tangent to guide the motion of the robot end-effector to perform surface following.}
The limitation of this strategy lies in the high demand on the accuracy of sensing, learning and control, as it depends strongly on the accuracy of end-effector positions and surface normal. 
{In addition, it lacks adaptability to different surfaces using force index to reflect the desired contact status, e.g., the desired force needs to be adjusted according to the feature of the surfaces to avoid the stick-slip phenomenon in \cite{liu2015finger}.}
In our work, we use the motion based control strategy. 
{Different from existing works, we use contact area rather than contact force to indicate the contact status, and guide the robot motion by directly mapping the sensors information to robot actions without computing the surface tangent.}
Furthermore, to the authors' best knowledge, this work is the first to directly map the tactile data to the robot motion for surface following, which can avoid accumulative error that exists in other works.

\subsection{Sensors used for Surface Following}
The common sensors used in surface following includes visual cameras {\cite{olsson2004flexible}}, force sensors {\cite{liu2015finger}} and tactile sensors {\cite{ohka2009object}}. 
{Visual cameras can be used to estimate the position of the target object, which is then used in combined vision/force control for interaction with a stiff uncalibrated environment in surface following tasks \cite{olsson2004flexible}.}
In addition to the surface following task, visual cameras have also been used to measure the planar-contour in contour following task \cite{baeten2002hybrid,chang2004cartesian}. 
As the camera is located away from the object surface, vision could be occluded by the robot manipulator that limits its application in practice. As a manner of contact sensing, force sensors have been used in the surface following {since the 1990s} \cite{bossert1996experimental, nunes1994sensor}. {In recent years,} a 6-axis force/torque sensor, covered with a deformable rubber skin, was integrated onto the fingertip of a robotic finger to perform surface following of a computer mouse \cite{liu2015finger}. 
{An improved intrinsic contact sensing algorithm is proposed, which can provide accurate estimation of the contact location with the deformable finger skin even at high friction forces.} 
However, the control quality could be deteriorated due to the inaccurate estimation of surface normal under high friction and the error of finger position estimation caused by the accumulated sensing errors of the finger joint angles \cite{liu2015finger}.
In addition, the force sensor can only provide the information of a single contact point each time resulting in limited sensing ability. 
Compared to force sensors, tactile sensors can provide  distributed multi-point tactile information of the contact \cite{luo2017robotic, luo2016iterative,luo2019iclap}, localise the contact~\cite{luo2015localizing} and predict the shape~\cite{luonovel} and pose~\cite{bimbo2016hand} of the object in hand .
In \cite{ohka2009object}, an optical fiber based tactile array sensor is developed and a simple contour tracing task was performed but no learning was involved. 
In \cite{chebotar2014learning}, two dynamical matrix analog pressure sensors are equipped on a robot gripper to provide tactile feedback for a task of gently scraping a surface with a spatula. The deviations of actual tactile data and desired tactile trajectory is used to correct robot movements. This tactile feedback is added to the system through perceptual coupling and its parameters is optimized using reinforcement learning.
In this work, the sensor has no direct contact with the surface, therefore, detailed contact information was not able to be acquired. In addition, due to the large size (16cm$\times$16cm) and low spatial resolution (20mm), it is not suitable to be equipped onto robot end-effector for surface following. In this paper, we use a high-resolution GelSight tactile sensor \cite{luo2018vitac,lee2019touching} to provide tactile feedback. With the detailed contact information in the tactile images, the motion of the manipulator end-effector can be facilitated to follow different object surfaces.

\section{Methodologies}
\label{sec:Methodologies}
As previous mentioned, we achieve end-to-end learning for surface following by direct mapping raw sensor data to robot actions 
draw from deep reinforcement learning by Google DeepMind \cite{mnih2015human}. 
The mapping is based on a novel proposed policy that exists in form of an artifical agent, termed surface following deep Q-network (SFDQN), which combines reinforcement learning with a deep neural network. 
The deep Q-learning algorithm used to train the artificial agent in \cite{mnih2015human} was applied to virtual game environment.
However, it faces trouble to be applied to real robot learning. For example, it may cause damage to sensor and robot due to continuous unreasonable actions according to output of SFDQN in the early training stage. To overcome this problem, we divide the standard training procedure into 2 steps: 1) generate the training and testing datasets on real robot using a designed behavior policy, and 2) train the SFDQN offline on computer using the obtained datasets. A noval index, $ContactRate$, is proposed to characterize the contact area and it is also used for a reward function when training the SFDQN. In this section, we describe the methods in detail, including the definition of $ContactRate$ and image processing, the elements of reinforcement learning, the designed behavior policy and generated datasets, and the model architecture and training algorithm for SFDQN.

\subsection{ Contact Rate and Image Processing}
\label{subsec:ImageProcessing}
 We observed through experiments that the change region of the GelSight sensor image increase monotonously as the contact between the sensor and the object surface increases in the initial contact stage. It indicates that maintaining contact area in a certain range is a good way to achieve good surface following performance. The intuitive idea to characterize contact area is to count the non-zero-pixel ratio of the subtraction image between the contact and non-contact images, which we called $ContactRate$. 

In our experiments, the GelSight sensor image is not only used for calculation of $ContactRate$, but also as partial input of SFDQN. To make it work more efficiently, we do image processing so as to: (1) remove extra information to speed up the data flow; (2) extract contact status information to calculate the reward for the deep Q-learning. 

\begin{figure}
 	\centering
 	\begin{subfigure}[b]{0.23\textwidth}
 		\centering
 		\includegraphics[width=3.5cm]{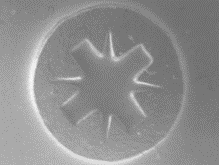}
 		\caption{}
 		\label{Fig:OnContactImage}
 	\end{subfigure}%
 	\hfill
 	\begin{subfigure}[b]{0.23\textwidth}
 		\centering
 		\includegraphics[width=3.5cm]{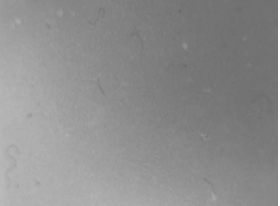}
 		\caption{}
 		\label{Fig:GelSightBackground}
 	\end{subfigure}%
 	 	\hfill
 	\begin{subfigure}[b]{0.23\textwidth}
 		\centering
 		\includegraphics[width=3.5cm]{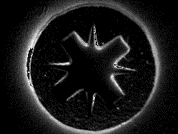}
 		\caption{}
 		\label{Fig:ObjectOutline}
 	\end{subfigure}%
 	 	\hfill
 	\begin{subfigure}[b]{0.23\textwidth}
 		\centering
 		\includegraphics[width=3.5cm]{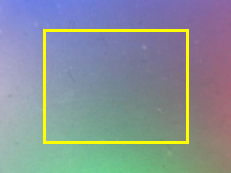}
 		\caption{}
 		\label{Fig:sensitiveZone}
 	\end{subfigure}%

 	\caption{Illustration of outlining the object in contact. (a) GelSight sensor feedback image of a contact object. (b) GelSight sensor background. (c) The outline of the target object. (d) Sensitive zone of GelSight Sensor.} 
 	\label{Fig:image process}
 \end{figure}

The GelSight sensor outputs  colored images of a resolution of $640\times480$. Prior to feeding these images into the 
SFDQN, we first remove the color information and resize the image to lower resolution ($64\times48$) (as shown in Fig. \ref{Fig:OnContactImage}). After that, we store the non-contact image (when the sensor is not in contact with the object surface) as the image background (as shown in Fig. \ref{Fig:GelSightBackground}). 
By subtracting the background from the contact image, we obtain the subtraction image. As the image is sensitive, we use a threshold filter to remove noisy pixels. Then we can see the outline of the contact area (as shown in Fig. \ref{Fig:ObjectOutline}). Finally, we use Eq. (\ref{eq:contactrate}) to calculate the $ContactRate$.

\begin{equation}\label{eq:contactrate}
    ContactRate = \frac{number\  of\  none\ zero\  pixels}{total\  pixel\  number} \times  1000
\end{equation}

\subsection{State, Action and Reward}
As a surface following problem in reinforcement learning (RL) model, the state should include information of the robot end effector, the target surface, their relative position and velocity. The states of the robot arm, i.e., the angular and velocity values of each joint, can be acquired. The information of the surface and the relative position can also be read from the GelSight sensor feedback. 
We combine the feedback image from sensor and the joints as the RL state as well as the input of 
SFDQN.

\begin{figure}
    \centering
	\includegraphics[width=8cm]{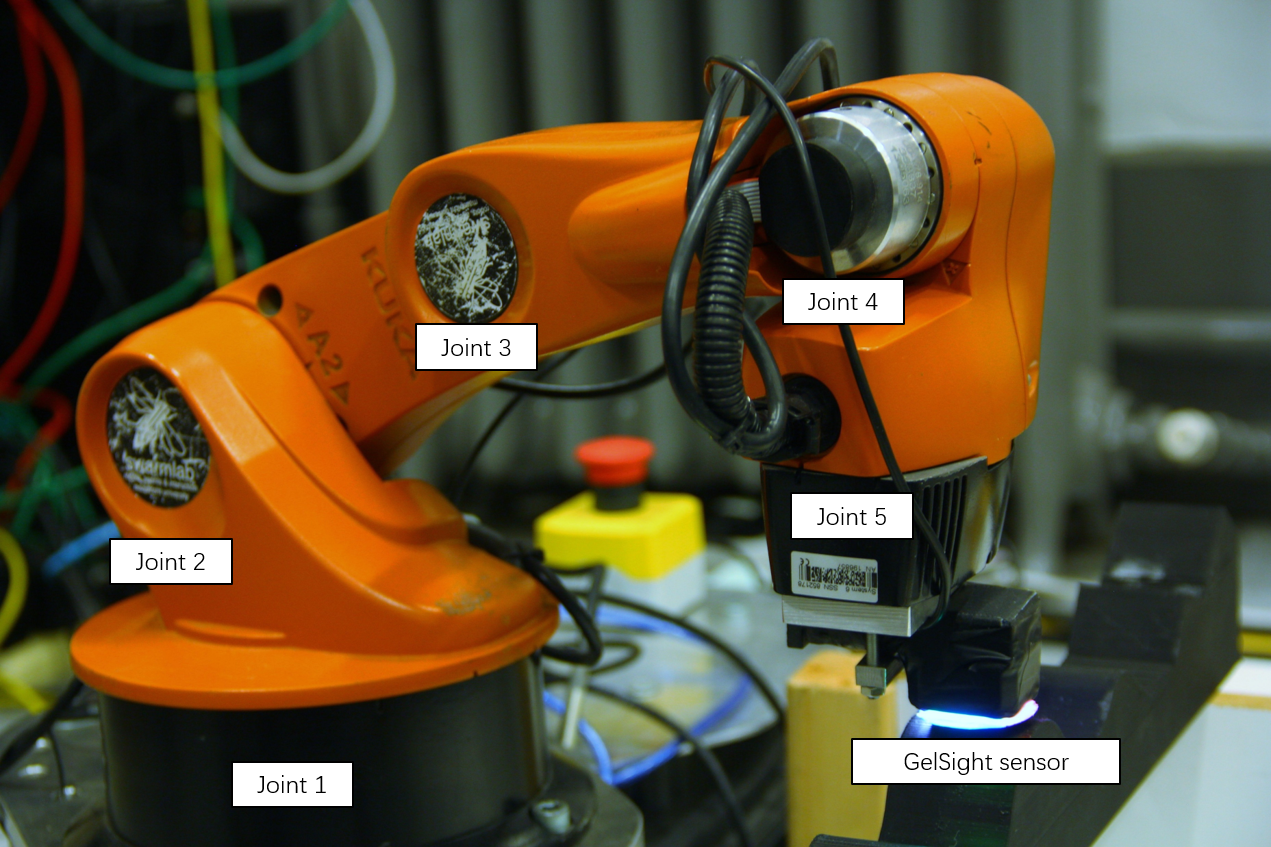}
	\caption{Training the youBot arm: its 3rd and 4th joints are under the control using the deep Q-learning program.} 
	\label{Fig:YouBot}
\end{figure}

\begin{figure}
    \centering
	\includegraphics[width=8cm]{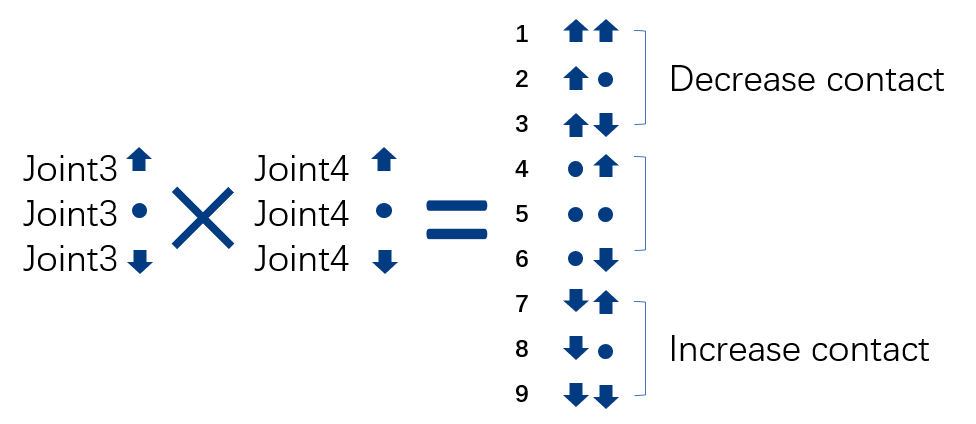}
	\caption{Definition of 9 actions corresponding to 2 joints.} 
	\label{Fig:actionDefine}
\end{figure}

The RL actions should match up with the actual movements performed by the robot. We define actions based on the joints motion. Each joint can move forward, backward and remain still. For $n$ joints, the number of actions can be $3^n$. We define these actions as the output of SFDQN. In this paper, we use 2 joints to simplify the action design. The 3rd and 4th joints are programmed to execute the actions as shown in Fig.~\ref{Fig:YouBot} and 9 actions can be generated as shown in Fig.~\ref{Fig:actionDefine}. Additionally, the angular shift of a single action (0.2 rad by default) can be adjusted by an auxiliary program.

The calculation of reward is based on the $ContactRate$ using Eq. (\ref{eq:reward}).
If the sensor-surface contact after the action's execution is in the desired contact status
($Contact Rate$ in the range of $[cr\_min, cr\_max]$), the agent will receive a reward 10, otherwise there will be no reward. 

\begin{equation}\label{eq:reward}
reward = \left\{ \begin{array}{cl}
10, & i \in [cr\_min, cr\_max],\\
0, & otherwise.
\end{array} \right.
\end{equation}

\subsection{Behavior Policy and Generated Datasets}
\label{subsec:bahavior policy}
Generating dataset is a necessary and important task prior to training of SFDQN.
Given a state $s$, the standard algorithm in  \cite{mnih2015human} select an action using $\epsilon$-greedy strategy, that is, with probability $\epsilon$ select a random action $a$, otherwise select the action with maximum $Q$ value according to SFDQN output.
In our problem, an ideal training dataset might be generated in a real surface following scenario. 
However, creating such dataset needs the GelSight sensor surface always rub on the object surface. In the early training stage, the actions selected according to SFDQN output is unreasonable and it may cause damage to the reflective membrane of the GelSight sensor or even the robot.  
Instead of pursuing the perfect dataset, 
an independent behavior policy is designed to generate as many contact statuses as possible, which is presented in Algorithm \ref{algorithm:behavior policy}. 

\begin{algorithm}[!htbp]
\SetAlgoLined
 \SetAlgoCaptionSeparator{.}
\caption{Dataset generation algorithm using behavior policy} \label{algorithm:behavior policy}
Initialize dataset $D$ as an empty set;\\
Initialize the desired number of units in $D$ to N; \\
Initialize the desired range of $Contact Rate$ to $[cr\_min, cr\_max]$, and set regional median value $cr\_ideal$ to $(cr\_min+cr\_max)/2$;\\
Initialize the set of actions $A$, and its subsets $A\_ic$, $A\_dc$, $A\_uc$ in which are the actions that can increase, decrease, unchange the $Contact Rate$, respectively;\\

\For {$units\_num = 1$ \KwTo $N$} {
    $\bullet$ Set the state $s$ with GelSight sensor image and robot joints value (position and velocity);\\
    \uIf{$units\_num \% 10 \geq 5$}{
        $\diamond$ Randomly select an action $a \in A$;\\
    }  
    \Else{
        $\diamond$ Calculate $cr$ according to s using Eq. (\ref{eq:contactrate});\\
        \uIf{$cr \geq cr\_ideal$}{
            Randomly select an action $a \in A \backslash A\_ic$;\\
            }
        \Else{
            Randomly select an action $a \in A \backslash A\_dc$;\\
        }
    }
    $\bullet$ Execute action $a$ on robot and update state $s'$;\\
    $\bullet$ Calculate reward $r$ according to state $s'$ using Eq. (\ref{eq:reward});\\ 
    $\bullet$ Store unit $\langle s,a,r,s' \rangle$ in $D$; \\
}
\end{algorithm}

At each time step, the behavior policy generates an action $a$ to map the current state $s$. 
The designed behavior policy contains 2 kinds of rules. One is complete random rule that randomly select an action from all the actions to enrich the diversity of dataset. The other is partial random rule that randomly select an action from a subset of actions which can drive the robot to the desired contact status. The latter can increase the ratio of positive reward in the dataset which can benefit the training of SFDQN. It also avoids cumulative actions that leeds to high $ContactRate$, which can damage the GelSight sensor.

To implement the partial random rule, we classify the actions into 3 subsets according to their effects on the change of $ContactRate$: 'increase', 'decrease' and 'unchange' the $ContactRate$. The classification can be easily done by tests, or even by analysis when it is not complicated as shown in Fig. \ref{Fig:actionDefine}.
The rule is as follows: if the current $ContactRate$ is greater or equal to the median value of the desired contact status range, we randomly select an action from the subsets other than the 'increase' subset; otherwise, we select an action from the subsets other than the 'decrease' subset.

In our experiment, the behavior policy uses 2 rules alternatively. It generates actions for 5 states using partial random rule and for the following 5 states using complete random rule, and then back to partial random rule. This is repeated until we get enough data for training of SFDQN. During the process, we record the state $s$, the generated action $a$, the new state $s'$ after $a$ is executed, and the reward $r$ calculated based on the GelSight sensor image regarding to $s'$ using Eq. (\ref{eq:contactrate}). The generated dataset is composed of units in form of $\langle s,a,r,s' \rangle$.


\subsection{Model architecture and training algorithm for SFDQN}
We use an artificial agent, SFDQN, for mapping the current information (state) to the robot motion (action) draw from the excellent ability of learning policies directly from hign-dimensional sensory inputs using end-to-end reinforcement learning. 
The goal of SFDQN is to find an optimal policy that offers the best action for given state, so as to maximize the discounted future reward. 
As the GelSight sensor that used to collect information from object surface generates sequences of high-resolution images as feedback, we design SFDQN using a deep convolutional neural network, which is especially good at extracting information from raw images,
to approximate the optimal action-value (also known as $Q$) function
\begin{equation}\label{eq:q_star}
	\begin{split}
        Q^\star (s, a) = \underset{\pi}{\max} \mathbb{E} [R_t + \gamma R_{t+1} + \gamma^2 R_{t+2} + \cdots\\
        | S_t = s, A_t = a, \pi]
	\end{split}
\end{equation}
which is the maximum sum of rewards $R_t$ discounted by $\gamma$ at each time step, 
achievable after taking an action $a$ under state $s$ according to a behavior policy $\pi$.
As $Q^\star$ function obeys the Bellman equation, Eq. (\ref{eq:q_star}) can be rewritten as
\begin{equation}\label{eq:Q_star_bellman}
	\begin{split}
       Q^\star(s,a)=  \mathbb{E}_{s'}[r+\gamma \underset{a'}{\max} Q^\star(s',a'|s,a]
	\end{split}
\end{equation}
where $r$ denotes the reward after taking action $a$, $s'$ and $a'$ denotes the state and action in the time step next to $s$ and $a$, respectively. 
For training of the Q-network, we define the loss function as follows:
\begin{equation}\label{eq:loss function}
	\begin{split}
       L_i(\theta_i)\ = \mathbb{E}_{ (s,a,r,s')} [ ( r+\gamma \underset{a'}{\max} Q(s',a';\theta^-_i)\\
       -Q(s,a;\theta_i) )^2 ]\\
	\end{split}
\end{equation}
where $\gamma$ is a discount factor, $\theta_i$ denotes the weights of the Q-network at iteration $i$, and $\theta^-_i$ denotes the weights of another Q-network used to compute the target at iteration $i$.

\begin{figure*}
    \centering
	\includegraphics[width=16cm]{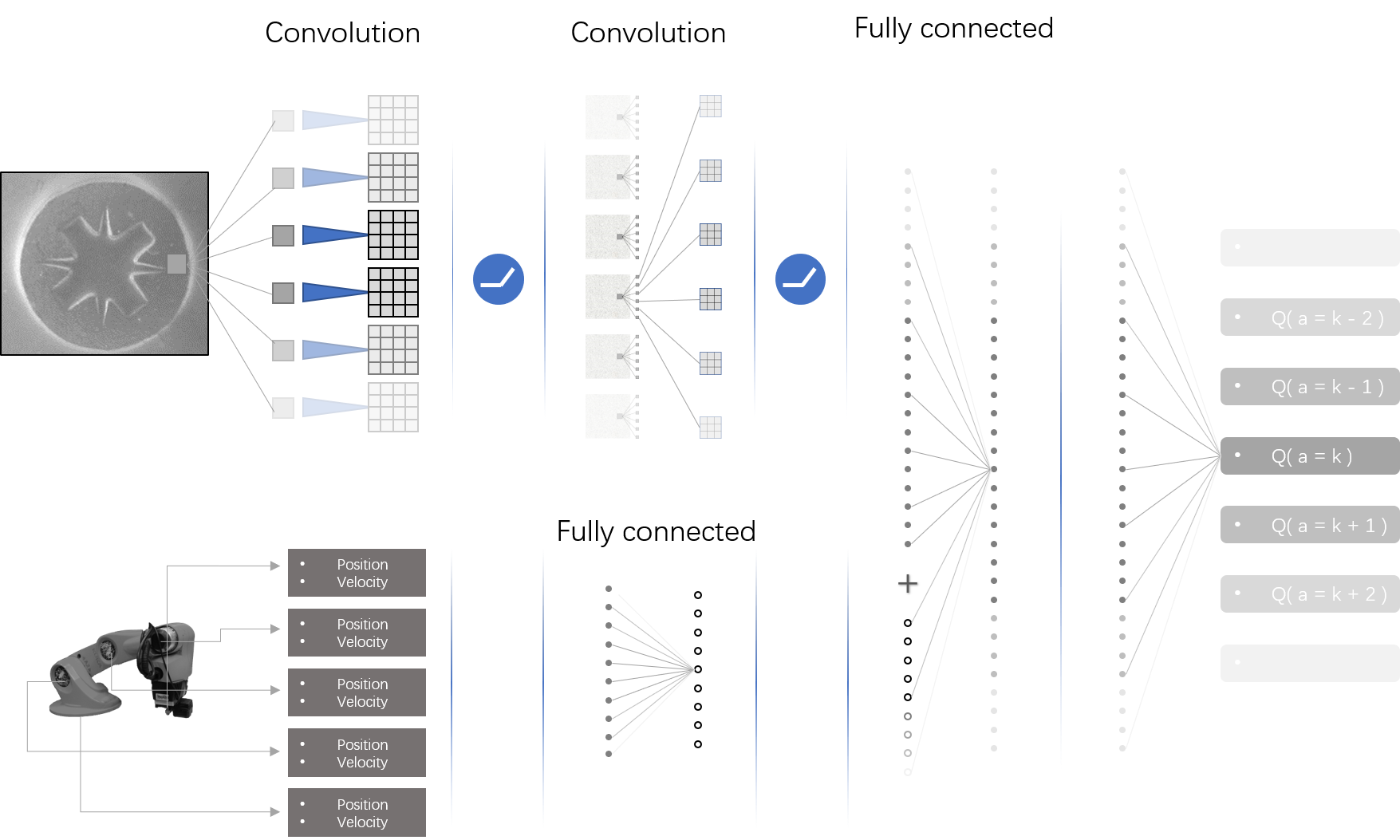}
    \caption{Schematic illustration of the Q-network for surface following deep Q-network (SFDQN). Given a state as input, the Q-network output the $Q$ value for each valid action. The input to the neural network consists of preprocessed GelSight sensor image and robot joint position and velocity. The image input is followed by several convolutional layers, while the joint information input is followed by a fully connected layer. Then the processed data flows are merged using a concatenate layer and followed by 2 fully connected layers with a single output, $Q$ value, for each valid action. Each convolution layer is followed by a ReLU activation function.}
	\label{Fig:CNNstructure}
\end{figure*}

Fig.~\ref{Fig:CNNstructure} shows the Q-network that used to parameterize an approximate value function $Q(s,a;\theta_i)$.
Given a state as input, the Q-network output the $Q$ value for each valid action.
The preprocessed GelSight sensor image (as mentioned in Section \ref{subsec:ImageProcessing}) is accepted as the main input to the Q-network, and followed by several convolutional layers to extract the feature information.
The position and velocity values of robot joints are fed to the Q-network as auxiliary input, and followed by a fully connected layer (dense layer). 
Then the processed data flow from two inputs are merged using a concatenate layer.
And then after 2 following dense layers, the Q-network will output $Q$ value for each action. 

To obtain the best approximation of optimal $Q^\star$ function, we use deep Q-learning algorithm to train the Q-network $Q$ for SFDQN. 
We apply $unit~replay$ and $iterative~update$ to overcome the problems associated with neural network type function approximator. 
The former randomly samples $T$ units as training data at each time step, which can prevent correlated input to the Q-network. 
The later uses another Q-network $\hat{Q}$, which only updates the weight every $C$ steps, to calculate the Q-learning target: $r + \gamma \max_{a'} Q(s',a';\theta^-_i)$. 
The reason of doing this is that if the Q-learning target is calculated by the main Q-network $Q$, every weight update will also change the label data distribution and this could make the learning unstable.
The detailed algorithm is presented in Algorithm \ref{algorithm:deep Q-learning}.

\begin{algorithm}[!htbp]
\SetAlgoLined
 \SetAlgoCaptionSeparator{.}
\caption{Deep Q-learning algorithm for training of Q-network for SFDQN} \label{algorithm:deep Q-learning}
Initialize action-value function $Q$ with random weights $\theta$; \\
Initialize target action-value function $\hat{Q}$ with weights $\theta^- = \theta$; \\
Generate training dataset $D$ according to behavior policy using Algorithm \ref{algorithm:behavior policy}, and divide $D$ into two datasets, $D\_train$ and $D\_test$, which are used for training and testing, respectively;\\
\For {$timeStep = 1$ \KwTo $M$} 
{
    $\bullet$ Randomly sample $T$ units $\langle s,a,r,s' \rangle$ from $D\_train$; \\
    $\bullet$ \For {$t = 1$ \KwTo $T$}
    {
        $\diamond$ Input $s$ to $Q$ and set $y = Q(s,a;\theta)$; \\
        $\diamond$ Input s' to $\hat{Q}$ and set $\hat{y} = max_{a'} \hat{Q}(s',a';\theta^-)$; \\
        $\diamond$ Compute loss fuction value $L = (r + \gamma \hat{y} - y)^2$ and perform a gradient descent step on it with respect to the network parameters $\theta$; \\
    }    
    $\bullet$ Every $C$ steps reset $\hat{Q}$ to Q by setting $\theta^- = \theta$;\\
    $\bullet$ Every $E$ steps record $\theta$ for evaluation of $SFDQN$ on testing dataset $D\_test$;
}
\end{algorithm}

\section{Experiment Setup}
\label{sec:ExperimentSetup}

The experiment was conducted on a KUKA youBot platform, with a GelSight tactile sensor as the end effector. In order to connect and control KUKA youBot and GelSight tactile sensor and to perform neural network computation, several ROS nodes are designed to deal with the data stream in learning and testing stage. OpenCV library is employed to assist GelSight sensor image process. The neural network computation is GPU accelerated.


Finally, when the learned model is ready to be assessed, we fix its weights and modify it to be an surface following action generator.

\subsection{Components of the experiment platform}
YouBot is a mobile manipulator platform\footnote{KUKA youBot platform: http://www.youbot-store.com/}, developed for the purpose of basic level robotics education, cognitive-manipulation research and industrial-oriented application development. 
The two main components of the youBot are the 5 DOF arm and the mobile platform. The youBot operation command can be assigned to 9 joints, 5 located on the arm, 4 on the mobile platform. As a result, the youBot arm and the mobile platform can each carry out an independent task simultaneously. The original end-effector - a gripper, is replaced with the GelSight sensor. The youBot has an internal computer 
but of relatively weak computation power, therefore, it is necessary to run the deep Q-learning program on a external computer.


The key innovation of GelSight sensor is the use of inward reflective membrane that can adapt to various textures. A group of LED unit is fixed inside the sensor as illuminator. The camera is located in the center bottom of the sensor. Unlike traditional tactile sensors, GelSight sensor generates stream of high resolution images from a target object. More details of the sensor can be found in \cite{luo2018vitac,lee2019touching}. 

\subsection{ROS nodes}
Robot Operating System (ROS)\footnote{ROS: http://www.ros.org/} has a collection of tools specialized for robot tasks. ROS manages a complex robot manipulation task by turn its subtasks into ROS nodes that effectively communicate between each other. In this experiment, nodes with following functions are created:
(1) read in keyboard command
(2) read in raw GelSight sensor data and process with $OpenCV$
(3) manage action command and transfer to youBot driver
(4) record and read observation data units
(5) create and train the deep neural network using Q-learning rule
(6) read in GelSight sensor data, generate action under behavior policy
(7) read in GelSight sensor data, generate action from SFDQN



\subsection{SFDQN setup}
In our experiment, we set $ContactRate$ in the range of 20 to 40 as the desired contact status.

In the training process, we use two SFDQN model. 
One model includes 2 convolutional layers to process the image input. The first CONV layer has 8 filter of size 4$\times$4, the second CONV layer has 16 3$\times$3 filters.
The other model is much deeper, and it uses 10 CONV layers with 5 4$\times$4 layers followed by 5 3$\times$3 layers.
The fully connected layers use default settings. 

The training of SFDQN is carried out using Algorithm \ref{algorithm:deep Q-learning}. The number of units in dataset $D$, denoted by $N$, is set to 12,000. 
The number of training steps $M$ is set to 20,000. 
The number of units sampled at each step, denoted by $T$, is set to 10. 
The synchronization interval of $Q$ and $\Hat{Q}$, denoted by $C$, is set to 500.
The interval of recording the weights of SFDQN, denoted by $E$, is set to 100.

\section{Experiments Results and Analysis}
\label{sec:ExperimentsResultsandAnalysis}

In this section, we will analyze the performance of proposed surface following approach, including the performance of the trained deep neural network model, the behavior policy used to generate various state-action pairs, and the image processing method.

\subsection{The performance of behavior policy}
As mentioned in Section \ref{sec:Methodologies}, we use a behavior policy to create various observation units as training data.

The advantage is obvious, it is easy to design, theoretically safe, moreover, the two rules that form up the behavior policy both generate random actions, as a result, a large enough dataset should cover most possible state transition near the sensor-surface contact point.

However, this policy has certain defects. First, it is has to be manually adjusted to fit actions to different arm poses, for example, if the youBot arm is moved to the opposite side of the mobile platform, the effect of carrying out action 8 is also inverted. Second, this policy is static, so it does not evolve like a neural network.

In the actual test, dataset generated by this policy did not contains high-contact states(contact rate > 300), which means the policy is safe enough to avoid risky situation.

\begin{figure}
    \centering
	\includegraphics[width=9cm]{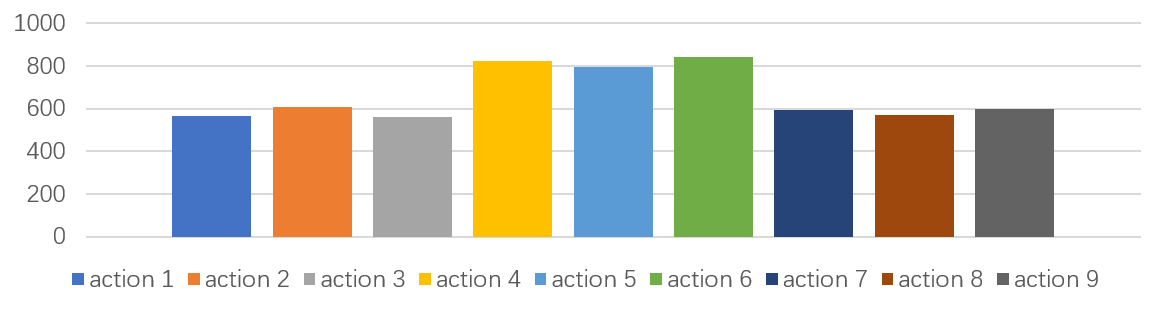}
	\caption{ The distribution of actions generated according to behavior policy.} 
	\label{Fig:actionDistribution}
\end{figure}

Fig.~\ref{Fig:actionDistribution} shows the distribution of actions generated according to behavior policy. We can see the behavior policy is able to create even action distribution, which helps to cover the state-action space more evenly so as to enhance the exploration effect in training of SFDQN.

\subsection{The Performance of SFDQN}

A trained SFDQN is expected to extract contact information from the GelSight sensor input, then generate actions that lead to desired $Contact Rate$. 

This ability is evaluated by checking if the action given by the SFDQN is in the list of $good$ actions, while $good$ action is determined by the $Contact Rate$ of input state. For example, when the input state has a low $Contact Rate$, actions that increase $Contact Rate$ are $good$, when the $Contact Rate$ is already in the desired region, actions that have trival effect of $Contact Rate$ are considered $good$. 

We input all state in a test dataset to the SFDQN and check the proportion of $good$ actions it generated. The proportion, also refered to as $pre$ision ,indicate sthe abilcity of a certain SFDQN model.
During the training, we also save the weight of the SFDQN periodically and evaluate all saved models using above method, this will give us the learning curve of a SFDQN with particular structure.

\begin{figure}
    \centering
 	\begin{subfigure}{0.5\textwidth}
 	    \centering
 		\includegraphics[width=9cm]{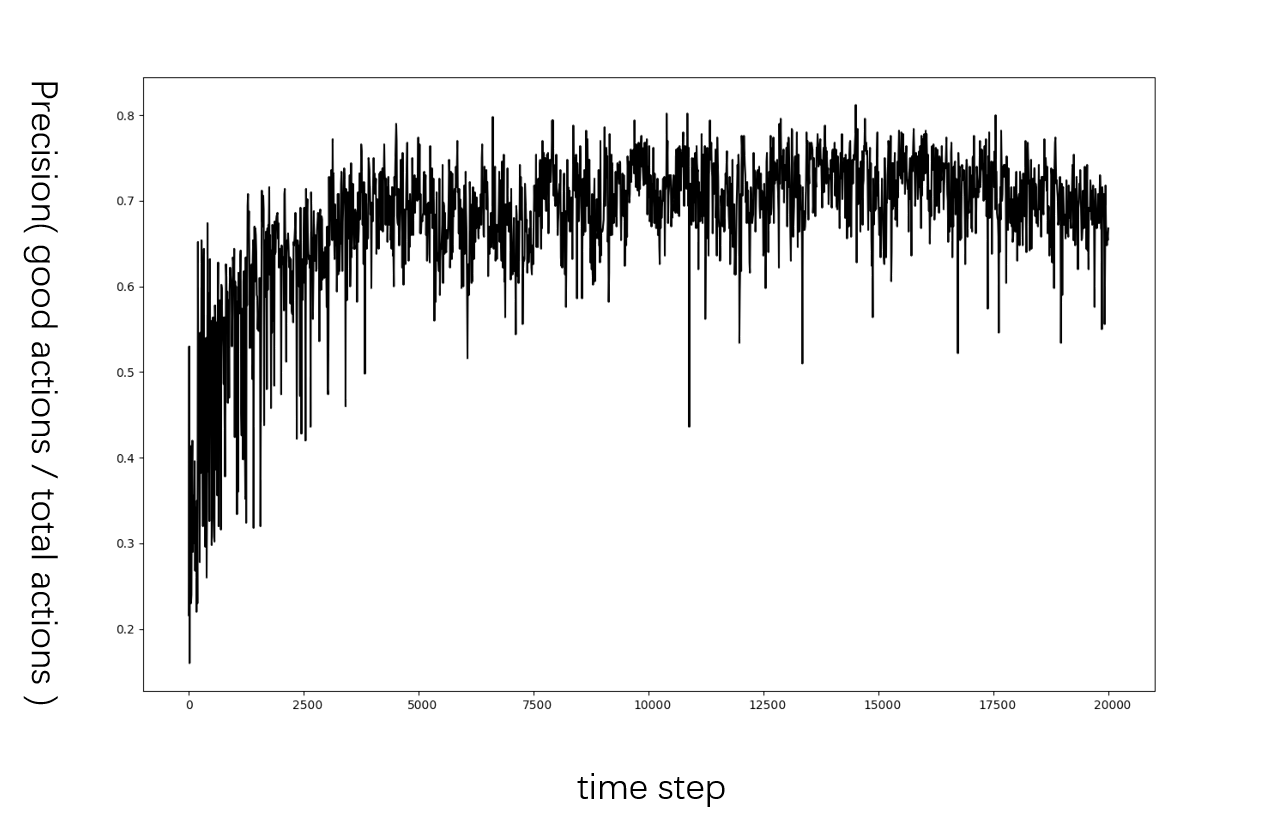}
 		\caption{}
 		\label{Fig:2CONVLayerNN}
 	\end{subfigure}
 	\hfill
 	
	\begin{subfigure}[b]{0.5\textwidth}
	    \centering 		\includegraphics[width=9cm]{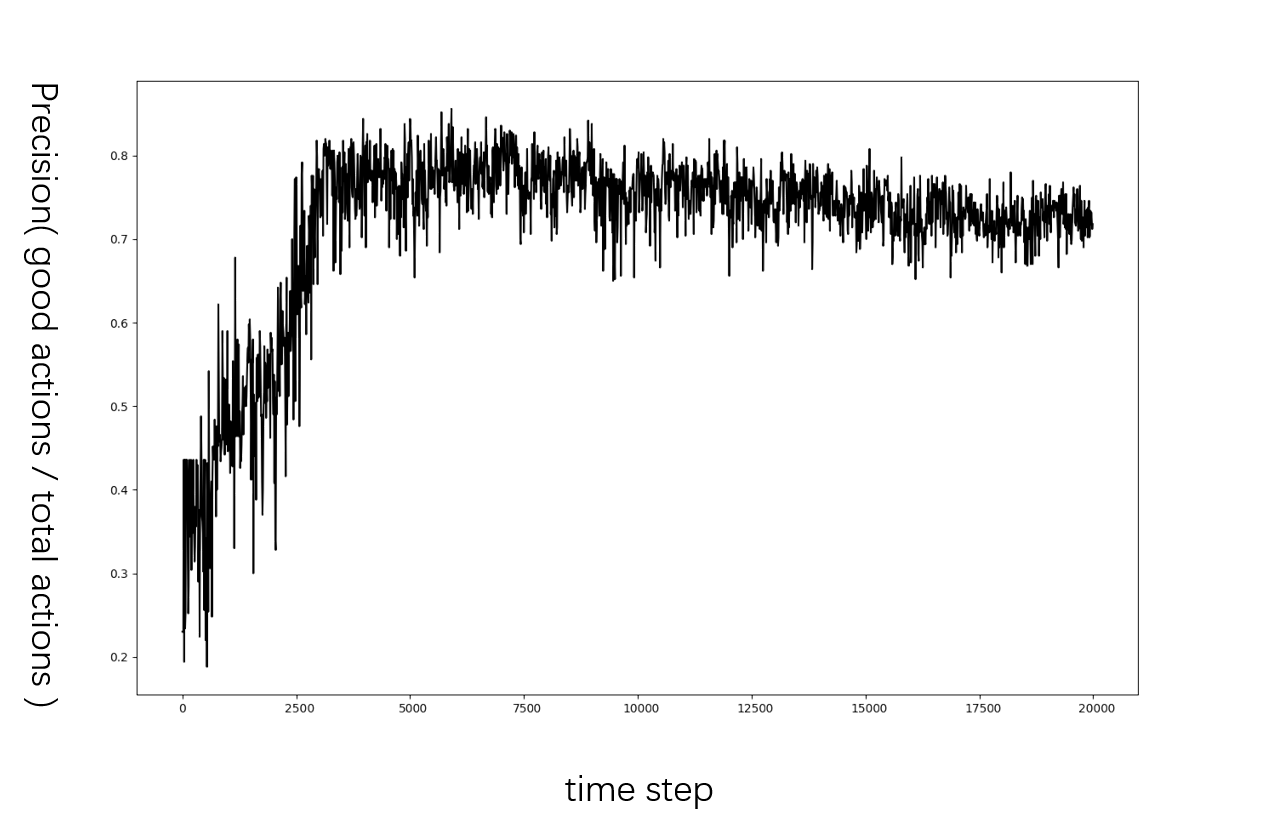}
 		\caption{}
 		\label{Fig:10CONVLayerNN}
 	\end{subfigure}
    \caption{Learning curve for the precision of action prediction. (a) Learning curve of a neural network model with 2 CONV layers. (b) Learning curve of a neural network model with 10 CONV layers.}
    \label{Fig:learningCurve}
\end{figure}

Fig.~\ref{Fig:learningCurve} shows the learning curve of two SFDQN models with 2 and 10 CONV layers, respectively. The horizontal axis shows the time step and the vertical axis shows the precision of action prediction. The initial learning rate is set to 0.0001. We can see that both models reached their performance peak after training for 5,000 time steps, and the deeper model tend to produce more stable action prediction compared to the shallow one.

\subsection{Real Surface Test}
After fully trained, we tested the neural network model as a surface following action generator on a real wood surface (Fig.~\ref{Fig:GelSightDoingSurfaceFollowing}). In the test, we move the youBot alongside the wood surface manually, the neural network model receives the changing contact rate and send the predicted action to the youBot arm. The whole system is able to perform the surface following action properly although the GelSight sensor left the test surface occasionally. We also tested the system by touching the sensor with finger while moving the finger up and down, it turns out that surface following action has a velocity upper bound, the arm cannot adapt to fast changing surface. The reason is, the current effect of an action is to update the joint angular value, the maximum speed is determined by number of predicted actions per second and it is possible to define an action to update the joint velocity, but velocity type command tend to have unstable delay time on youBot platform compare to angular position type command, which could make the training process more challenging.

\subsection{Discussion}
As shown in fig.~\ref{Fig:image process}, the proposed image process works well with GelSight sensor, the outline of the target object is clearly displayed. In addition, the method we used to calculate the contact rate has shown stable performance and is very sensitive to slight and medium level contact especially in the central area of the sensor. The main drawback of this method is that the calculation of contact rate relies on an independent background image. As the reward is defined based on contact rate, this means the whole training process will be influenced if the background is not correct. Once the reflective membrane on the GelSight sensor is relocated, we have to retrain the neural network to make it functional again. Another problem exists when using GelSight sensor on complex surface shape, e.g., surface with high slope. The edge area of the sensor is less sensitive the center area,  contact on the edge tend to be underestimated.


\section{Conclusion and Future Works}
\label{sec:Conclusions}
In this paper, we propose a novel surface following approach based on deep Q-learning algorithm using a GelSight sensor. We built up a experiment platform with a GelSight tactile sensor and a KUKA youBot. We ran a set of experiments to check the performance of the proposed solution. In conclusion, our proposed solution has reached more than 80\% of the theoretical maximum performance. The future research can be conducted as following: (1) To analyze different RL elements definition, e.g., take a sequence of GelSight sensor image as RL state, map input state to continuous action space, change action command type from position update to velocity update. (2) To extend the current solution to other surface following problems, e.g., contour following and surface exploration.

\section*{Acknowledgment}

Dr Jing Wang was supported by the China Scholarship Council for 1 year academic visit at the University of Liverpool.

{\small
	\bibliographystyle{ieeetr}
	\bibliography{root.bib}
}

\end{document}